\documentclass[conference]{IEEEtran}
\usepackage{cite}
\usepackage{amsmath,amssymb,amsfonts}
\usepackage{algorithmicx}
\usepackage{graphicx}
\usepackage{textcomp}
\usepackage{xcolor}

\usepackage{times}
\usepackage{helvet}
\usepackage{courier}
\usepackage{url}
\usepackage{graphicx}
\usepackage{amsmath}
\usepackage{algorithm}
\usepackage{booktabs}
\usepackage{tabularx}
\usepackage{hyperref}
\usepackage[noend]{algpseudocode}

\def\BibTeX{{\rm B\kern-.05em{\sc i\kern-.025em b}\kern-.08em
    T\kern-.1667em\lower.7ex\hbox{E}\kern-.125emX}}
\begin{document}

\title{AI solutions for drafting in Magic: the Gathering}

\author{\IEEEauthorblockN{Henry N. Ward}
\IEEEauthorblockA{\textit{Bioinformatics and Computational} \\
\textit{Biology Graduate Program} \\
\textit{University of Minnesota} \\
Minneapolis, Minnesota \\
wardx596@umn.edu} \\
\IEEEauthorblockN{Bobby Mills}
\IEEEauthorblockA{\textit{Draftsim.com} \\
San Diego, California \\
bobbyam115@gmail.com}
\and
\IEEEauthorblockN{Daniel J. Brooks}
\IEEEauthorblockA{\textit{Draftsim.com} \\
San Diego, California \\
daniel.brooks@caltech.edu} \\
\\
\IEEEauthorblockN{Arseny S. Khakhalin}
\IEEEauthorblockA{\textit{Biology Program} \\
\textit{Bard College}\\
Annandale-on-Hudson, New York \\
khakhalin@bard.edu}
\and
\IEEEauthorblockN{Dan Troha}
\IEEEauthorblockA{\textit{Draftsim.com} \\
San Diego, California \\
dan@draftsim.com}
}

\maketitle

\begin{abstract}
Drafting in Magic the Gathering is a sub-game within a larger trading card game, where several players progressively build decks by picking cards from a common pool. Drafting poses an interesting problem for game and AI research due to its large search space, mechanical complexity, multiplayer nature, and hidden information. Despite this, drafting remains understudied, in part due to a lack of high-quality, public datasets. To rectify this problem, we present a dataset of over 100,000 simulated, anonymized human drafts collected from Draftsim.com. We also propose four diverse strategies for drafting agents, including a primitive heuristic agent, an expert-tuned complex heuristic agent, a Naive Bayes agent, and a deep neural network agent. We benchmark their ability to emulate human drafting, and show that the deep neural network agent outperforms other agents, while the Naive Bayes and expert-tuned agents outperform simple heuristics. We analyze the accuracy of AI agents across the timeline of a draft, and describe unique strengths and weaknesses for each approach. This work helps to identify next steps in the creation of humanlike drafting agents, and can serve as a benchmark for the next generation of drafting bots.
\end{abstract}


\section{Introduction}
AI agents have recently achieved superhuman performance in several challenging games such as chess, shogi, go, and poker \cite{Silver2017MasteringKnowledge,Silver2018ASelf-play,Brown2019SuperhumanPokerc}, as well real-time strategy games such as StarCraft II and multiplayer online battle arenas (MOBAs) such as Dota 2 \cite{Vinyals2019GrandmasterLearning,OpenAI2019DotaLearning}. These successes open opportunities to branch out and to create game-playing AI for other complex games. Much like strategy and MOBA games, collectible card games (CCGs) such as Hearthstone and Magic: the Gathering (MtG) present challenging milestones for AI, due to their mechanical complexity, multiplayer nature, and large amount of hidden information \cite{Hoover2019TheHearthstone}. Although some studies investigate AI for Hearthstone \cite{Stiegler2016HearthstoneSystem,Garcia-Sanchez2018AutomatedHearthstone,Garcia-Sanchez2016EvolutionaryHearthstone,Swiechowski2018ImprovingAlgorithms}, relatively little work exists on building game-playing AI for MtG \cite{Johann2017DeckbuildingAlgorithm,Ward2009MonteGathering}.

In this paper, we focus on a game mode known as ``drafting'' that involves progressive deck-building, where players take turns to select cards for their collection from given initial sets of cards \cite{Kowalski2020EvolutionaryGenes}. From their final collections, each player builds a deck and plays games against each other to determine the winner of the draft. We focus on drafting in Magic: the Gathering. MtG features one of the most complicated and popular drafting environments, where eight players each open a pack of 15 semi-random cards, select one card from that pack, and pass the remainder of the pack to an adjacent player. This process is repeated until all cards are drafted, and then is repeated twice over for two additional packs (with the second pack passed in the opposite direction). By the end of the draft, each player possesses a pool of 45 cards from which they select 20-25 cards to build a deck. Critically, each player's current pick and growing collection is hidden from other players.

While the core idea of building a deck by progressively selecting cards is shared by all drafting games, other key aspects of drafting are variable. On the simpler end of the spectrum, Hearthstone's drafting game - known as Arena - presents a single player with three semi-random cards from which they choose one to add to their deck. While drafting in MtG and Hearthstone both involve progressive deck-building from randomly-assorted cards, unlike Hearthstone, MtG drafting is complicated by its competitive multiplayer nature, larger search space, and substantial amount of hidden information. It is therefore likely that successful MtG drafting agents will rely on strategies that generalize to other, simpler drafting games like Hearthstone Arena. A summary of key aspects in which popular drafting games vary is listed in Table 1.

\begin{table*}[t]
\centering
\caption{Properties of popular drafting games}
\begin{tabularx}{\textwidth}{@{\hskip 6pt\extracolsep{\stretch{1}}}*{6}{r}}
\toprule
Game & Drafting Mode & Hidden Info? & \# of Players? & Multiplayer? & Asynchronous? \\ \midrule
MTG & Draft & Yes & 8 & Yes & Both \\
Eternal & Draft & Yes & 12 & Yes & Yes \\
Hearthstone & Arena & No & 1 & No & Yes \\
Gwent & Arena & No & 1 & No & Yes \\
Legends of Runeterra & Arena & No & 1 & No & Yes \\
DotA & Team Formation & No & 10 & Yes & No \\
League of Legends & Team Formation & No & 10 & Yes & No \\
Autochess & Autobattler & Very little & 8 & Yes & No \\
Teamfight Tactics & Autobattler & Very little & 8 & Yes & No \\ \bottomrule
\end{tabularx}
\end{table*}

These attributes of drafting games present interesting challenges for AI agents, and are further complicated by the large search space of a single draft. When evaluating a single pack of cards to decide which card to pick, a successful MtG drafting agent must take into account the individual strength of each card, the synergies between each card and the cards already in the player's collection, and the strategies their opponents may be pursuing. In a full draft, a drafting agent sees a total of 315 cards and makes 45 consecutive decisions of which card to take, bringing the total number of direct card-to-card comparisons to 7,080. In addition, the agent could potentially compare the distribution of cards it receives from its opponents to an expected distribution, as well as note which cards were not drafted after each pack of 15 cards traveled around the table of 8 players for the first time. Overall, for a set of 250 different cards, the full landscape of a draft comprises about 10\textsuperscript{700} starting conditions and roughly 10\textsuperscript{40} potential deck-building trajectories, rendering brute-force approaches infeasible.

A central goal for AI drafting agents is, rather than to draft optimally, to emulate human drafting. As online CCGs such as MtG Arena frequently pit players against drafting bots and not against other players, the creation of human-like agents could benefit both players and game development teams. While drafting bots currently employed in games such as MtG Arena enable asynchronous play, they seem to rely on simple heuristics that draft predictably and can result in negative gameplay experiences for players \cite{Henke2019ArenaInsight,Henke2019IsDecks}. Human-like drafting bots could provide a better experience and allow human players to build decks with more varied strategies. 

Here, we design, implement, and compare several strategies for building human-like MtG drafting agents. Because we do not aim to construct agents that draft optimally and because deck-building algorithms for completed drafts are beyond the scope of this work, we do not evaluate the quality of decks drafted by agents. Instead, we directly evaluate agents' ability to emulate human drafting strategies by measuring their predictions of human choices in a dataset of simulated MtG drafts performed by anonymous users of the website draftsim.com. We benchmark several drafting strategies, including a neural-network approach and a Bayesian approach, against simple heuristic-based approaches, and show that they predict human choices with relatively high accuracy.

Our work contributes the following:

\begin{itemize}
    \item We present the first large-scale public dataset of human MtG drafts, enabling the study of drafting for the AI and game development communities. This is downloadable at \href{http://draftsim.com/draft-data}{draftsim.com/draft-data}. 
    \item We frame drafting as a classification problem with the goal of creating drafting agents that accurately predict human card choices, and perform detailed evaluations of drafting agents to identify their relative strengths and weaknesses in addition to measuring their overall accuracy.
    \item We show that a deep neural-network approach best predicts human card choices, and suggest that similar approaches can help game developers interested in building new drafting agents.
\end{itemize}

\section{Relevant Work}

Several papers investigated constructed deck-building in Hearthstone and MtG \cite{Garcia-Sanchez2016EvolutionaryHearthstone,Johann2017DeckbuildingAlgorithm,MesentierSilva2019EvolvingMeta,Chen2018Q-DeckRec:Games,Hoover2019TheHearthstone}. These works framed deck-construction as a combinatorial optimization problem, where the goal is to select cards to add to a deck that maximize the win rate of the completed deck against a set of known decks. Evolutionary algorithms, Q-learning, and a utility system were employed to address this problem. Similar work applied evolutionary or quality-diversity algorithms towards deck-building in Hearthstone and Dominion with the aim of investigating or evolving game balance \cite{Bhatt2018ExploringSpace,Mahlmann2012EvolvingDominion,Fontaine2019MappingBoundaries}. However, these approaches are difficult to apply to MtG drafting \cite{Johann2017DeckbuildingAlgorithm}, due to the inherent variability of the game and the large amount of hidden information involved. Moreover, the scarcity of public data from actual drafts makes the evaluation of drafting agents challenging.

Unfortunately, previous research on constructed deck-building cannot be directly reapplied to building human-like drafting bots. To leverage this body of work, assumptions would have to be made about decks to optimize against before the draft is completed. This is a potentially difficult problem in its own right: even disregarding the high variability of drafts within a specific format, players only select roughly half of the cards they draft to put into their deck. Assuming that players use exactly 17 basic lands in their deck and thus choose 23 cards from their total pool of 45 cards to complete a 40-card deck, there are well over 10\textsuperscript{12} possible ways to construct a deck from a single draft. Accordingly, a crucial prerequisite for framing drafting as a combinatorial optimization problem is the existence of an efficient algorithm for building a realistic deck from a completed draft pool, which is beyond the scope of this work. 
 
Outside of online CCGs, a small body of research examines drafting in multiplayer online battle arenas (MOBAs) such as League of Legends and DotA 2 \cite{LucasHanke2017AGames,Chen2018TheDrafting,Summerville2016Draft-AnalysisLearning,Yu2019E-SportsNetwork}. MOBAs task players with sequentially choosing a team of in-game avatars that have different skillsets and synergies with each other, alternating choices with the opposing team which they aim to beat during a subsequent game. Much like drafting in MtG, MOBA drafting involves sequentially choosing heroes from a depleting pool. Previous research on this topic aimed to build recommender systems for hero choices \cite{LucasHanke2017AGames,Chen2018TheDrafting}, or to predict hero choices as drafts progress \cite{Summerville2016Draft-AnalysisLearning,Yu2019E-SportsNetwork}. Like the work presented here, these authors typically use machine learning techniques to predict the next chosen hero or the optimal combination of heroes based on training datasets of actual drafts.

Drafting games seen in MOBAs as well as the autobattler genre likely benefit from existing research on team formation problems. Previous work has formulated fantasy football as a sequentially-optimal team formation problem \cite{Matthews2012CompetingDomains}. The key distinctions between selecting players in fantasy football and selecting heroes during MOBA drafts are the gameplay phases that occur after each fantasy football draft (MOBAs only have one gameplay phase, which occurs after the draft is completed), the fact that MOBA heroes are limited resources (unlike fantasy football players, which can be chosen by multiple human players), and the economy system in fantasy football that assigns prices to players (MOBA drafting lacks an economy system). Autobattlers are perhaps more similar to fantasy football, because they also feature an economy system and alternate drafting phases and gameplay phases. However, heroes selected in autobattlers, like heroes in MOBA drafts, are a limited resource. While it may be possible to build drafting agents for MOBAs and autobattlers that treat the games as team formation problems, further research is necessary to enable the adaptation of previous team formation work to these games or to MtG drafting \cite{Datta2012CapacitatedNetworks,Gutierrez2016TheSociometry,Lappas2009FindingNetworks}. 

Although little previous work exists on drafting in online CCGs, a recent study applied evolutionary algorithms towards arena drafting in a CCG named Legends of Code and Magic \cite{Kowalski2020EvolutionaryGenes}. While the evolutionary strategy applied may be applicable towards other arena drafting modes, in addition to the aforementioned issues adapting constructed deck-building work to MtG drafting, the authors' simplifying assumption that card choices do not depend on prior choices is likely inappropriate for a synergy-driven game like MtG (discussed below). Other existing work on drafting in online CCGs consists of an early offshoot of this project that since developed independently \cite{Saxe2020BotDraftsim}, as well as several other websites that allow users to perform simulated drafting against bots \cite{CardsphereCardsphere:Simulator,MTGDraftKingMTGDraftKing:Gathering}. 

\section{MtG Terminology}

Here we describe terminology for drafting games in general, and for MtG in particular.

\subsection{Drafting games}

We define drafting games as games which include drafting as a core mechanic. Drafting is defined as sequential deck-building from a rotating, limited resource. Drafting games typically include one or more drafting phases, where players build or improve their decks, and gameplay phases, where players pit their deck against other players’ decks. These phases are nearly always discrete. 

\subsection{MtG overview}

MtG (Magic: the Gathering) is a card game in which two or more players use their decks to compete against each other in one or more matches. A player wins by reducing their opponent's life total from the starting value of 20 to 0. Each player starts with a hand of seven cards and draws one additional card each turn. Players use ``land'' cards to generate mana, which can then be used to play other cards. Mana comes in five different colors (white, blue, black, red, and green), and is used to cast ``spell'' cards at the cost of some combination of different colors of mana. While most cards are limited to four copies per deck, a deck can have an unlimited amount of special lands known as ``basic lands.''

\subsection{Domain complexity}

MtG's comprehensive rules are approximately 200 pages long \cite{WizardsoftheCoast2013Magic:Rules}, and judges are employed at every significant tournament to resolve frequent rule-based misunderstandings. Moreover, previous work has shown that the set of MtG rules is Turing-complete \cite{Churchill2019Magic:Complete}, and the problem of checking whether or not a single MtG action is legal can be coNP \cite{Chatterjee2016TheGathering}.  

In total, there are over 17,000 MtG cards. However, a given set used for drafting typically contains somewhere between 250 and 350 unique cards. On occasion, drafts may include multiple packs from different sets.

\subsection{Card features}

Magic cards are distinguished by a variety of categorical features. Each card belongs to one or more of the seven major types: creature, sorcery, instant, artifact, enchantment, land, planeswalker, and tribal. Each non-land card has a mana cost, which represents the amount of mana required to play it. This mana cost may include both colored mana and colorless mana (which can be of any color).

\section{Data description and exploratory analyses}

\subsection{Data description}

The website draftsim.com offers a simulated drafting experience, and has collected anonymized draft data from its users from spring 2017 onwards. Draftsim allows human players to draft against 7 bots that each follow the same manually tuned heuristic strategy, which is described in detail below (the ``complex heuristic'' bot). At the beginning of a draft, the user picks a set of cards to draft from. Each set typically contains around 250-300 different cards and is specifically balanced around drafting. From the chosen set, booster packs are randomly generated with a card frequency distribution matching that of real booster packs. In total, 11/15 cards are ``common'' rarity, 3/15 are ``uncommon'', 7/120 are ``rare'', and 1/120 are ``mythic.'' After drafting, the user's choices as well as choices made by the 7 bots they draft against are recorded anonymously and saved in an SQL database. Draftsim does not record player IDs, their IP or location, or the time it took them to make individual picks. As of summer 2020, Draftsim holds data for all sets released after Jan 2018 (14 at the time of writing), with about 100,000 completed drafts for popular sets. Each full draft (one data point) consists of 45 consecutive decisions made by one player.

\subsection{Training and testing data}

All drafting agents presented below were trained and evaluated on drafts of a single MtG set, Core Set 2019 (abbreviated as M19). This set contains 265 mechanically different cards: 16 mythic rare cards, 53 rare cards, 80 uncommon cards, 111 common cards, and 5 basic lands. We obtained a total of 107,949 user M19 drafts and split these into a training set of 86,359 drafts and a testing set of 21,590 drafts (an 80/20 split). As each draft contains 24 unique packs of cards, a total of 2,072,616 packs were seen by drafting agents during training, and a total of 518,160 packs were used to evaluate the drafting agents’ performances. 

\subsection{Analysis and visualization of the data}

Cards of every MtG set form a complex network of interactions. Some cards are designed to ``synergize'' with each other, either because they require similar resources to be played (such as mana of the same color), or because one card changes its behavior when another card is present in the game. Certain pairs of cards can also complement or compete with each other in terms of optimizing the distribution of mana costs within a deck. The goal of drafting is thus typically not to pick the ``objectively best'' card from a pack, as drafting instead resembles a stochastic descent towards powerful deck configurations where players try to pick cards that are both individually strong and synergize with each other. These deck configurations are explicitly created in each set by the game’s developers and are unique for each drafting environment.

To inform the creation of AI agents and to visualize this multidimensional optimization landscape, for each pair of cards $(i,j)$ in our data we looked into whether they were likely to be drafted together by human players. We calculated the frequency of pairs of cards ending up in the same collection $P_{ij}$, and compared it to the probability that these cards would be drafted together by chance if drafting was completely random ($P_i P_j$). Two cards may be called synergistic if $P_{ij}>P_i P_j$, and the ratio of two probabilities $S_{ij} = P_{ij}/P_i P_j$ can be used as a measure of this synergy. 

To visualize synergies within each set, we considered a weighted graph, defined by a distance matrix $D$, that placed cards closer to each other if they were often drafted together, with card-to-card distances given by a formula: $D_{ij} = (1-S_{ij}/\text{max}_{ij}(S_{ij}))$. We projected this graph to a 2D plane using a linear multidimensional scaling algorithm that maximized Pearson correlation coefficients between the ``true'' distances $D_{ij}$ and the Euclidean distances within the 2D projection $\hat{D}_{ij}$. Visual profiles, or ``synergy plots,'' for several MtG sets are shown in Fig. 1, with each card represented by a point and the color of each point matching the color of mana required to play that card.

\begin{figure*}[!tbp]
  \begin{minipage}[b]{1\textwidth}
    \includegraphics[width=1\textwidth]{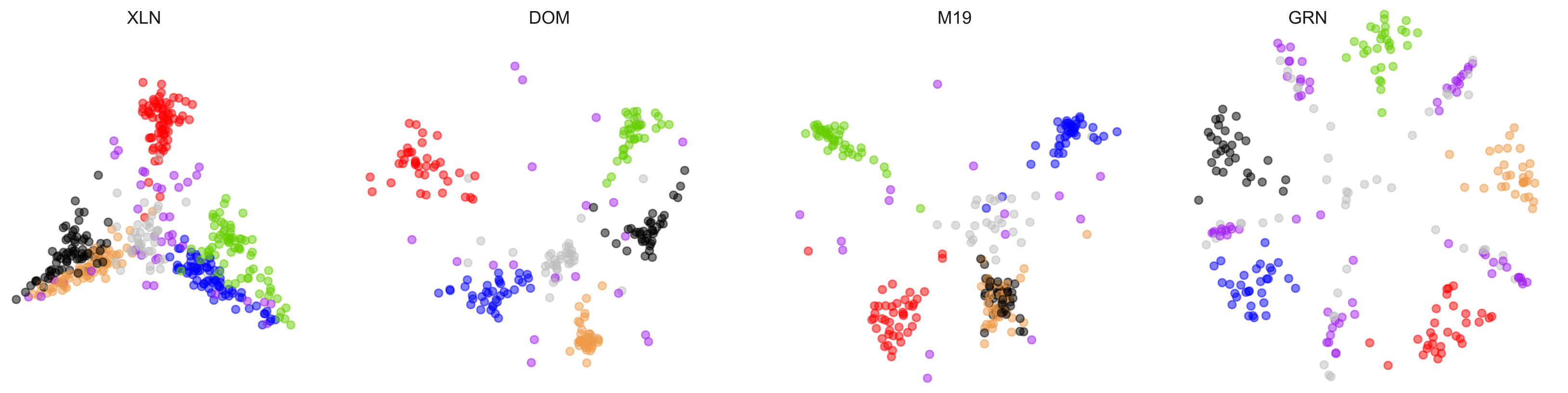}
    \caption{Synergy plots for Draftsim data collected from Ixalan (XLN), Dominaria (DOM), Core Set 2019 (M19), and Guilds of Ravnica (GRN).}
  \end{minipage}
\end{figure*}

Synergy plots also illustrate why a good drafting AI algorithm is crucial for making virtual drafting compelling for human players. It is clear from the synergy plots that cards indeed form clusters based on whether users believe they can be used together in a viable deck. For M19 these clusters are primarily defined by card colors, whereas XLN  was designed to emphasize relatively color-independent strategies as a so-called ``tribal set'' (Fig. 1) \cite{Rosewater2020ToGATHERING}. As eight players engage in a draft, they are competing for synergistic decks, and are each trying to settle into deck archetypes represented by one of these clusters. As a result of this competition, final decks can be expected to be most streamlined and powerful if each player commits to a certain archetype, freeing cards that belong to other archetypes to other players. This interplay between competitive and collaborating drafting explains the demand for difficult-to-exploit drafting bots that is expressed by players in online discussion venues \cite{Henke2019ArenaInsight,Henke2019IsDecks}.

\section{Drafting Agents}

\subsection{Goals for drafting agents}

Using the data described above, we designed drafting agents that approximate the behavior of human players. In other words, we trained and evaluated bots on their ability to predict the card that was actually picked by a human player at each stage of a draft, given the choice of cards in the pack and cards already in the player’s collection. 

\section{Agents}

We constructed five different drafting agents which implement different drafting strategies. Two of these agents, \emph{RandomBot} and \emph{RaredraftBot}, serve as baselines to compare other bots against. They implement, respectively, random card-picking and card-ranking based on simple heuristics. \emph{DraftsimBot} ranks cards based on heuristic estimations of their strength and whether their color matches the color of cards already in the agent’s collection. \emph{BayesBot} ranks cards based on estimated log-likelihoods of human users picking cards from a given pack, given the current collection. \emph{NNetBot} ranks cards based on the output of a deep neural network trained to predict human choices (Table 2). The agents are described in detail below.

\subsection{RandomBot: random drafting}

This baseline agent ranks all cards in a pack randomly.

\subsection{RaredraftBot: simple heuristics}

As a more realistic baseline, RaredraftBot emulates a drafting strategy used by inexperienced human players by drafting the rarest card in each pack. It breaks ties between cards of the same rarity by randomly choosing between all cards whose color matches the most common color in the bot’s collection. 

\subsection{DraftsimBot: complex heuristics}

The drafting algorithm currently employed on draftsim.com and implemented here ranks cards based on approximations of individual card strength provided by a human expert, as well as on whether each card matches the most common colors among the stronger cards in the bot’s collection. These strength ratings are set in the 0-5 range by a human annotator, with higher numbers assigned to stronger cards. The DraftsimBot agent aims to approximate the behavior of human drafters: it first chooses stronger cards based on these human-provided annotations, and as the draft progresses it develops an increasing tendency to draft cards ``in-color,'' until at some point it switches to a strict in-color drafting strategy.

The rating for a card $c$ is given by the scalar output of the function $rating$. The bot’s collection is represented as a vector of card names, $d$. Rating is computed as the sum of a strength rating, $strength(c)$, and a color bias term, $colorbias(c)|_d$.

\begin{equation}
rating(c)|_d = strength(c) + colorbias(c)|_d
\end{equation}

A card’s colored mana costs are expressed in the vector $colors$. The $i$th component of the 5-dimensional color vector represents the required number of mana of the $i$th color. 

\begin{equation}
colors(c)[i]=\textrm{required mana of ith color}
\end{equation}

The pull of a card, $pull(c)$, is the amount that its strength exceeds a threshold for minimum card strength (2.0). For M19, the weakest 20\% of the set (53 cards) does not meet this threshold. This excludes the weakest cards effects' on drafting.

\begin{equation}
pull(c)=max\big(0, strength(c) - 2.0\big)
\end{equation}

A player’s commitment to a color, $colorcommit$, is calculated by summing the pull of cards in that player’s deck $D$ containing that color. The color bonus, $C(i)$, is a heuristic computed from $colorcommit$ that balances picking powerful cards and committing to a two-color pair. This bonus is calculated differently early in the draft (the ``speculation'' phase) and later in the draft (the ``commitment'' phase).

\begin{equation}
colorcommit[i]|_D = \Sigma_{c\in D \ | \ colors(c)[i] > 0} \; pull(c)
\end{equation}
\begin{equation}
bonus(c) = max(0.257 * colorcommit[i], \; 0.9)
\end{equation}
\begin{equation}
bonus(c) = \Sigma (bonus(on) - bonus(off))-0.6
\end{equation}

The agent starts in the speculation phase, where one-color cards of color $i$ are assigned a bonus proportional to that player’s $colorcommit$ for this color and capped at 0.9, as shown in Equation (5). Colorless cards are assigned a bonus equal to the maximum color bonus of 1-color cards. Multicolored cards with 2-3 colors are assigned a rating equal to the bonus of the card's colors, subtracting off the bonus of other colors and a multicolored penalty of 0.6 as shown in Equation (6). Multicolored cards with 4+ colors are ignored by the model and assigned a bonus of 0. This phase lasts until the playing agent is committed to 2+ colors, or until the fourth pick of the second pack, whichever comes first.

The agent is considered committed to a color $i$ when the summary $colorcommit[i]$ exceeds 3.5, and it can be committed to 0, 1, or 2+ colors. If the agent is committed to 2+ colors, the two colors with the greatest $colorcommit$ values are referred to as that agent’s primary colors. During the committed phase, on-color cards are assigned a large bonus of +2.0, while off color cards are assigned a penalty of $-$1.0 for each off color mana symbol beyond the first. 

\subsection{BayesBot: Bayesian card-ranking}

This agent explicitly relies on the co-occurrence statistics of the training drafts, and gives priority to pairs of cards that were drafted together more frequently than by chance.

The BayesBot agent employs one strategy to pick the first card within a draft and another to pick all other cards. For the first card, it looks for the card that was most often picked first by human players in the first pack. In practice, for all picks made by human players in the training set, it counts cases $m_{ij}$ when both cards $i$ and $j$ were present in the pack and sub-cases $m_{i>j}$ when card $i$ was picked earlier than card $j$, to calculate the probability that card $i$ is picked over card $j$. Then it employs the Naive Bayes approximation, assuming independence of probabilities $P(i>j)$ and relying on pre-calculated log-likelihood values, to find the card with the highest probability of being picked first in Equation (8).

\begin{equation}
P(i>j) = m_{i>j}/m_{ij}
\end{equation}
\begin{equation}
i = \text{argmax}_i \prod_j P(i>j) = \text{argmax}_i \sum_j \log(m_{i>j}/m_{ij})
\end{equation}

For all subsequent cards, the agent maximizes the ``synergy'' between a newly picked card and cards already in the collection. In practice, it counts all cases $n_{ij}$ when card $i$ was present in the pack and card $j$ was already in the collection, and all sub-cases when after that the card $i$ was actually drafted, $n_{i\rightarrow j}$. The ratio of these two numbers gives a marginal probability that card $i$ is drafted when card $j$ is already collected in Equation (9).

\begin{equation}
P(i\rightarrow j\;|\;i\in pack \land j\in collection) = n_{i\rightarrow j}/n_{ij}
\end{equation}
\begin{equation}
P(i) = \prod_j P(i\rightarrow j | i,j) = \prod_j n_{i\rightarrow j}/n_{ij}
\end{equation}
\begin{equation}
i = \text{argmax}_i \prod_j n_{i\rightarrow j}/n_{ij} = \text{argmax}_i \sum_j \log(n_{i\rightarrow j}/n_{ij})
\end{equation}

In the Naive Bayes approximation, the probability of drafting a card $i$ given a full collection of cards $\{j\}$ is equal to a product of probabilities $P(i\rightarrow j|i,j)$ across all cards in $\{j\}$, as these probabilities are (naively) assumed to be independent. Therefore, the full probability $P(i)$ of drafting a card $i$ is assumed in Equation (10). The card with the highest total probability is drafted, and the top-ranked card is also calculated using log-likelihood values in Equation (11).

In practice, these formulas are vectorized as $P = Q \cdot c$, where $Q_{ij} = \log(n_{i\rightarrow j}/n_{ij})$ is a matrix of log-transformed probabilities of drafting card $i$ from the pack to join card $j$ already in the collection, and $c$ is the indicator vector of all cards in the collection ($c_j$ is the number of cards of type $j$ present in the collection). Note that while $Q$ seems to quantify ``attraction'' between the cards, it also indirectly encodes the rating of each card, as the values of $Q_{ij}$ are larger when the card $i$ is strong and the probability of it being drafted $P(i\rightarrow j)$ is high. 

\subsection{NNetBot: deep neural network card-ranking}

This agent applies a naive deep learning approach to emulate human drafting. From vector representations of the pack, the current collection, and each actual pick, the card picked by the human player is represented as the target variable $y$: a one-hot encoded vector of length $S$, where $S$ is the number of cards in the set ($S=265$).

The independent variable $x$ is constructed from a vector encoding the collection (with each element $x_i$ representing the number of copies of the card $i$ in the collection). This collection information $x$ of length $L$ is fed into the input layer of a network with 3 dense layers, each $L$ elements wide, with leakyReLu activation ($a$=0.01), batch normalization, and dropout ($p$=0.5). The final layer is linear and projects to a one-hot-encoded output vector $\hat y$ of length $S$.

We also defined a pack vector $[x_{pack}]$ to represent cards present in the current pack. The model did not use the pack vector as an input, but rather used the current collection to predict the best cards in the entire set that could be picked. The output of the model was then element-wise multiplied by the pack vector to enforce the constraint that only cards in the pack can be selected in Equation (12).

\begin{equation}
pick = \text{argmax}(\hat y \odot x_{pack})
\end{equation}

To evaluate the model’s performance on the training dataset, the network was trained for 20 epochs using cross entropy loss with 3-fold cross-validation. All cross-validation accuracies stabilized at roughly 64.5\% after a handful of epochs, indicating that 20 epochs was sufficient training time. After cross-validation, a model was trained for 20 epochs on the entire training dataset and similarly stabilized at 64.7\% accuracy on the testing set. This model’s card rankings were used as the output of the NNetBot. 

\section{Comparisons of bot performance}

To evaluate whether our agents drafted similarly to human players, we measured each drafting agent’s top-one accuracy for predicting actual user choices across all 21,590 drafts in the testing set (Table 2). While all complex agents outperformed the baseline Random and RaredraftBot agents, the deep neural-network agent NNetBot outperformed the DraftsimBot and BayesBot agents. All differences between groups for both top-one accuracy were significant (Tukey test $p$ $<$ 2e-16).

In addition to measuring the overall accuracy of each agent, we also measured their top-one accuracy for each turn of the draft (from 1 to 45). Because pack size decreases as packs are depleted, from 15 cards to 1 card for picks 1 to 15, 16 to 30 and 31 to 45, per-pick accuracy could better reflect the players’ shifting goals during the draft compared to overall accuracy. The first card of each pack, for example, is often picked based on its rarity or its strength. Cards in the middle of each pack are more likely to be picked based on their synergy with existing collections, while the last cards of each pack are typically not desirable to any player. Per-pick accuracy for all bots on the testing set is shown in Fig. 2.  

Per-pick accuracies further highlight the differences between each agent. First, the NNetBot consistently outperformed all other bots, while the DraftsimBot and BayesBot performed similarly. All three complex bots outperformed the two baseline bots. Second, all bots except for the RandomBot performed better at the beginning than during the middle of every pack. This supports the assumption that players’ goals shift throughout the draft: players are more likely to pick cards based on estimated strength or rarity early on in each pack, and are more likely to pick cards for other reasons (such as synergy with their collection) during the middle of each pack. Third, the RaredraftBot performed better than the RandomBot across all picks, showing that agents with simple heuristics make a far more compelling baseline for assessing drafting agents than a randomized agent. 

\begin{table}[t]
\centering
\caption{Implemented drafting agents}
\begin{tabularx}{0.48\textwidth}{*{4}{>{\centering\arraybackslash}X}}
\toprule
Agent & Type & Needs training? & Mean testing accuracy (\%) \\ \midrule
RandomBot & Random & No & 22.15 \\
RaredraftBot & Heuristic & No & 30.53 \\
DraftsimBot & Heuristic & No & 44.54 \\
BayesBot & Bayesian & Yes & 43.35 \\
NNetBot & Deep neural network & Yes & 48.67 \\ \bottomrule
\end{tabularx}
\end{table}

Lastly, we also compared each card’s estimated strength, based on the expert-provided ratings of each card used by the DraftsimBot, to each agent’s accuracy in picking that card across all drafts (Fig. 3). Fig. 3 shows that the average pick accuracy varied greatly across different card strengths and different drafting agents. The RandomBot only successfully picked weak cards, as these cards were the last to be drafted. The RaredraftBot and BayesBot agents accurately picked both weak cards and strong cards, but struggled to pick medium-strength cards. The DraftsimBot and especially the NNetBot outperformed all other agents for medium-strength cards, but performed slightly worse for weak and strong cards. 

\begin{figure}[!tbp]
  \centering
  \begin{minipage}[b]{0.48\textwidth}
    \includegraphics[width=\textwidth]{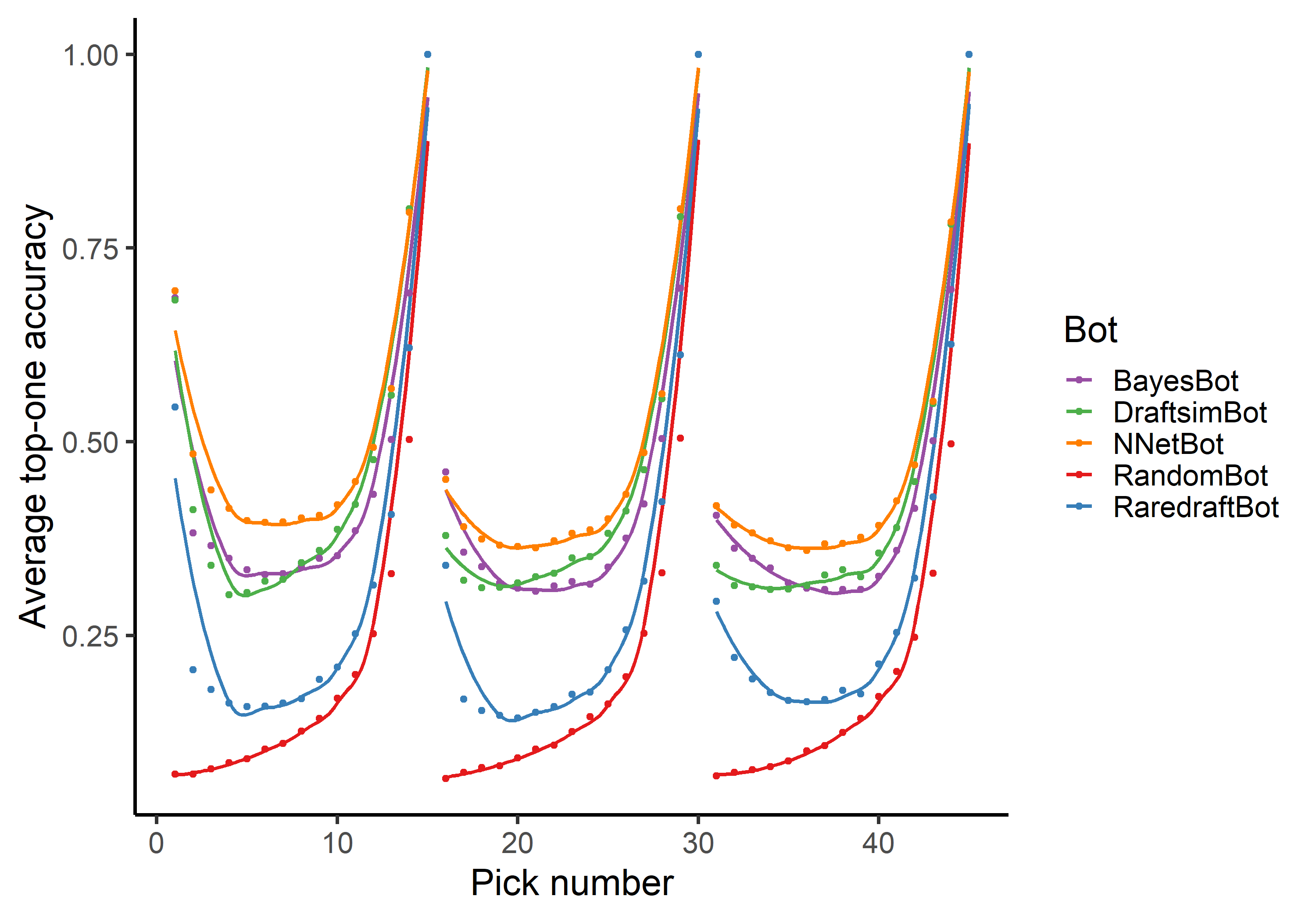}
    \caption{Per-pick accuracies for five bots.}
  \end{minipage}
  \hfill
  \begin{minipage}[b]{0.48\textwidth}
    \includegraphics[width=\textwidth]{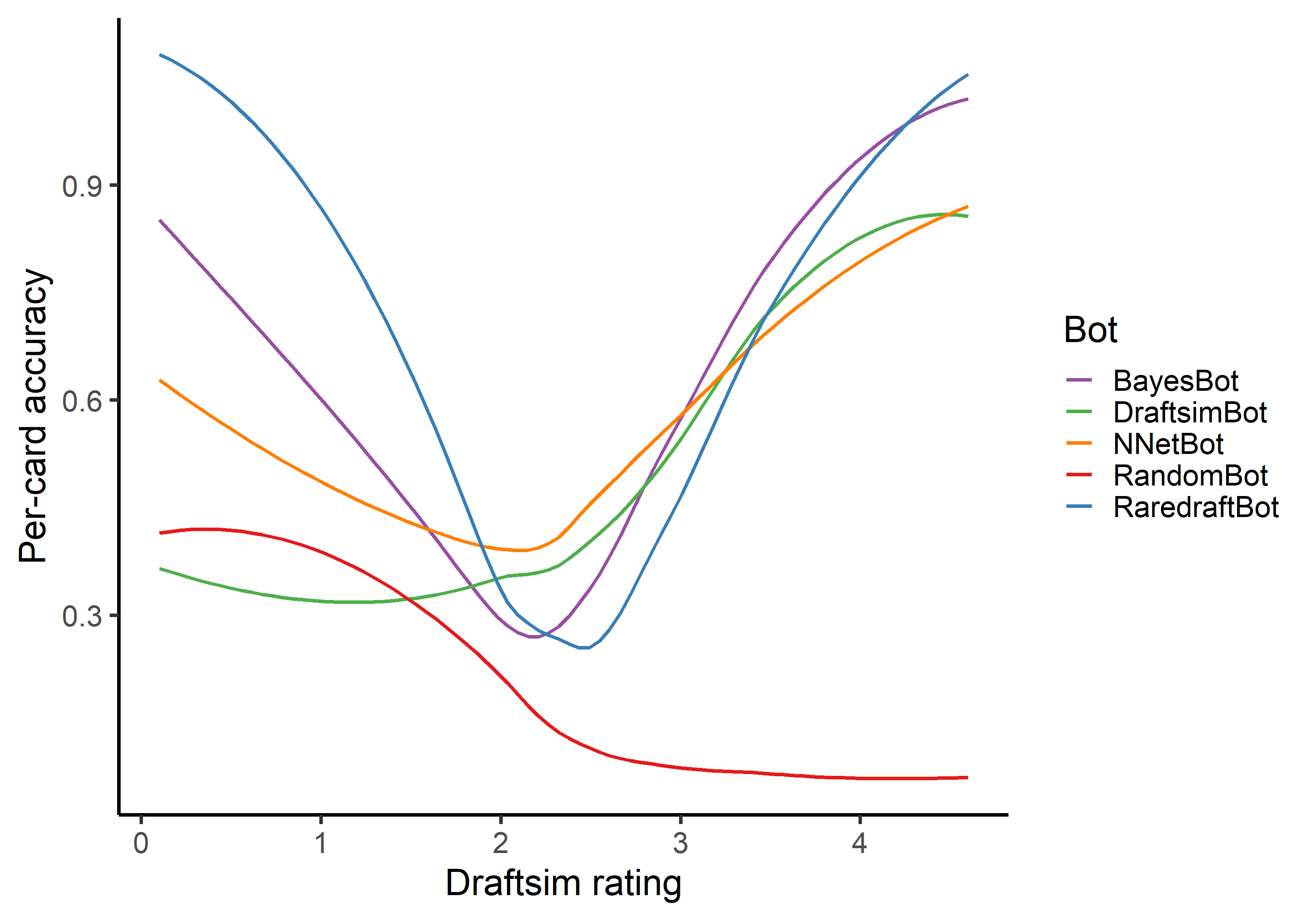}
    \caption{Pick accuracy vs. card strength for five bots.}
  \end{minipage}
\end{figure}

The most probable explanation for this surprising under-performance of the two best AI agents for edge-case cards lies in a social phenomenon known as ``raredrafting.'' A small share of cards in every set are designed to be weak for drafting environments, but are much sought after for other formats. In real drafts, players are tempted to take these weak but expensive cards instead of cards that could help them within the draft. While all drafts in the dataset were virtual, it seems that many players still raredrafted and picked weak rares instead of stronger common or uncommon cards as their first or second picks. This raredrafting behavior threw off the top performing NNBot and DraftsimBot agents but was recapitulated by the BayesianBot, that directly learned these signals, as well as the RaredraftBot (by definition).

\section{Discussion}

In this report, we present a large-scale dataset of human MtG drafts and compare several approaches to building human-like drafting agents for that set. We suggest that human-like drafting agents should be evaluated based on how well they approximate human choices in a set of testing drafts. We show that a neural network-based agent outperforms other agents, including a random agent, agents that implement simple or expert-tuned heuristics, and a naive Bayesian approach. 

Our approach has some limitations. First, as both the training and testing data were produced by humans drafting in a simulated environment, it is possible that our data does not approximate competitive MtG drafting. Some players might have randomly clicked through cards in a draft. Some might have ranked cards using the ``suggestions'' function that displays expert card ratings which are similar to those used by the DraftsimBot. Some might have drafted suboptimally on purpose, trying to explore the flexiblity of the set or to pick cards that represented only a single strategy. These limitations should not impact the bots’ ability to emulate human drafting, but could hamper the bots’ ability to generalize to competitive drafting environments. Second, because deck-building algorithms for MtG are outside the scope of this work, we do not evaluate the strength of decks built from completed drafts in simulated matches. This may also impact how well the agents generalize to competitive environments. Lastly, we performed all analyses on data from a single set. Future work could investigate how well bots perform on different MtG sets. 

One important contribution of our paper is the benchmarking of more advanced agents against heuristic-based bots. As for \cite{Kowalski2020EvolutionaryGenes}, who benchmarked their proposed evolutionary deck-building strategies for the card game Legends of Code and Magic against both random strategies and card rankings from top players, we observe a substantial performance benefit for deck-building based on simple heuristics compared to random chance. We hope that our agents, or similar neural network, Bayesian and heuristic-based agents may serve as useful benchmarks for future drafting and deck-building work. 

Our work also outlines several avenues for the development of future drafting bots. One is to develop agents that can generalize to diverse MtG sets, tolerate the inclusion of previously unseen cards, and work with mixtures of cards from different sets \cite{Saxe2020BotDraftsim}. Another opportunity is to create agents that address cooperation and competition between players during a draft. Human players are known to make guesses about what their neighbors might be drafting, based on the statistics of packs that are handed to them (a strategy known as ``signaling''), and use these guesses to inform their picks. Some players may also take cards that don't benefit them but might benefit their opponents (a strategy known as ``hate-drafting''). Neither of these behaviors is addressed by agents presented in this paper. Finally, future work should also tackle algorithms for building decks from completed drafts. This step is a prerequisite for testing the strength of draft pools in simulated MtG games, and thus is required to close the loop and enable reinforcement learning through self-play.

\section{Code and Data Availability}

The collection of M19 drafts is available under a CC BY 4.0 International license at \href{http://draftsim.com/draft-data}{draftsim.com/draft-data}. All code used to implement and test drafting bots is available at \href{https://github.com/khakhalin/MTG}{github.com/khakhalin/MTG}.

\bibliographystyle{ieeetr}
\bibliography{citations.bib}

\begin{thebibliography}{10}

\bibitem{Silver2017MasteringKnowledge}
D.~Silver, J.~Schrittwieser, K.~Simonyan, I.~Antonoglou, A.~Huang, A.~Guez,
  T.~Hubert, L.~Baker, M.~Lai, A.~Bolton, Y.~Chen, T.~Lillicrap, F.~Hui,
  L.~Sifre, G.~Van Den~Driessche, T.~Graepel, and D.~Hassabis, ``{Mastering the
  game of Go without human knowledge},'' {\em Nature}, vol.~550, pp.~354--359,
  10 2017.

\bibitem{Silver2018ASelf-play}
D.~Silver, T.~Hubert, J.~Schrittwieser, I.~Antonoglou, M.~Lai, A.~Guez,
  M.~Lanctot, L.~Sifre, D.~Kumaran, T.~Graepel, T.~Lillicrap, K.~Simonyan, and
  D.~Hassabis, ``{A general reinforcement learning algorithm that masters
  chess, shogi, and Go through self-play},'' {\em Science}, vol.~362,
  pp.~1140--1144, 12 2018.

\bibitem{Brown2019SuperhumanPokerc}
N.~Brown and T.~Sandholm, ``{Superhuman AI for multiplayer poker},'' {\em
  Science}, vol.~365, pp.~885--890, 8 2019.

\bibitem{Vinyals2019GrandmasterLearning}
O.~Vinyals, I.~Babuschkin, W.~M. Czarnecki, M.~Mathieu, A.~Dudzik, J.~Chung,
  D.~H. Choi, R.~Powell, T.~Ewalds, P.~Georgiev, J.~Oh, D.~Horgan, M.~Kroiss,
  I.~Danihelka, A.~Huang, L.~Sifre, T.~Cai, J.~P. Agapiou, M.~Jaderberg, A.~S.
  Vezhnevets, R.~Leblond, T.~Pohlen, V.~Dalibard, D.~Budden, Y.~Sulsky,
  J.~Molloy, T.~L. Paine, C.~Gulcehre, Z.~Wang, T.~Pfaff, Y.~Wu, R.~Ring,
  D.~Yogatama, D.~W{\"{u}}nsch, K.~McKinney, O.~Smith, T.~Schaul, T.~Lillicrap,
  K.~Kavukcuoglu, D.~Hassabis, C.~Apps, and D.~Silver, ``{Grandmaster level in
  StarCraft II using multi-agent reinforcement learning},'' {\em Nature},
  vol.~575, pp.~350--354, 11 2019.

\bibitem{OpenAI2019DotaLearning}
{OpenAI}, {:}, C.~Berner, G.~Brockman, B.~Chan, V.~Cheung, P.~D{\c{e}}biak,
  C.~Dennison, D.~Farhi, Q.~Fischer, S.~Hashme, C.~Hesse, R.~J{\'{o}}zefowicz,
  S.~Gray, C.~Olsson, J.~Pachocki, M.~Petrov, H.~P. d.~O. Pinto, J.~Raiman,
  T.~Salimans, J.~Schlatter, J.~Schneider, S.~Sidor, I.~Sutskever, J.~Tang,
  F.~Wolski, and S.~Zhang, ``{Dota 2 with Large Scale Deep Reinforcement
  Learning},'' {\em arXiv preprint arXiv:1912.06680}, 12 2019.

\bibitem{Hoover2019TheHearthstone}
A.~K. Hoover, J.~Togelius, S.~Lee, and F.~de~Mesentier~Silva, ``{The Many AI
  Challenges of Hearthstone},'' {\em KI - K{\"{u}}nstliche Intelligenz}, 9
  2019.

\bibitem{Stiegler2016HearthstoneSystem}
A.~Stiegler, C.~Messerschmidt, J.~Maucher, and K.~Dahal, ``{Hearthstone
  deck-construction with a utility system},'' in {\em 2016 10th International
  Conference on Software, Knowledge, Information Management {\&} Applications
  (SKIMA)}, pp.~21--28, IEEE, 2016.

\bibitem{Garcia-Sanchez2018AutomatedHearthstone}
P.~Garc{\'{i}}a-S{\'{a}}nchez, A.~Tonda, A.~M. Mora, G.~Squillero, and J.~J.
  Merelo, ``{Automated playtesting in collectible card games using evolutionary
  algorithms: A case study in hearthstone},'' {\em Knowledge-Based Systems},
  vol.~153, pp.~133--146, 8 2018.

\bibitem{Garcia-Sanchez2016EvolutionaryHearthstone}
P.~Garcia-Sanchez, A.~Tonda, G.~Squillero, A.~Mora, and J.~J. Merelo,
  ``{Evolutionary deckbuilding in hearthstone},'' in {\em 2016 IEEE Conference
  on Computational Intelligence and Games (CIG)}, pp.~1--8, IEEE, 9 2016.

\bibitem{Swiechowski2018ImprovingAlgorithms}
M.~Swiechowski, T.~Tajmajer, and A.~Janusz, ``{Improving Hearthstone AI by
  Combining MCTS and Supervised Learning Algorithms},'' in {\em IEEE Conference
  on Computatonal Intelligence and Games, CIG}, vol.~2018-August, IEEE Computer
  Society, 10 2018.

\bibitem{Johann2017DeckbuildingAlgorithm}
S.~Johann, B.~Knut, A.~Fludal, and A.~Kofod-Petersen, ``{Deckbuilding in Magic:
  The Gathering Using a Genetic Algorithm},'' {\em Master's Thesis, Norwegian
  University of Science and Technology (NTNU)}, 2017.

\bibitem{Ward2009MonteGathering}
C.~D. Ward and P.~I. Cowling, ``{Monte carlo search applied to card selection
  in magic: The gathering},'' in {\em CIG2009 - 2009 IEEE Symposium on
  Computational Intelligence and Games}, pp.~9--16, 2009.

\bibitem{Kowalski2020EvolutionaryGenes}
J.~Kowalski and R.~Miernik, ``{Evolutionary Approach to Collectible Card Game
  Arena Deckbuilding using Active Genes},'' in {\em IEEE Congress on
  Evolutionary Computation}, 7 2020.

\bibitem{Henke2019ArenaInsight}
T.~Henke, ``{Arena Exploits: Beating Bots Black, Green, and Blue | Cardmarket
  Insight},'' 2019.

\bibitem{Henke2019IsDecks}
T.~Henke, ``{Is it Possible to Exploit the Bots in Arena Draft? -
  ChannelFireball - Magic: The Gathering Strategy, Singles, Cards, Decks},''
  2019.

\bibitem{MesentierSilva2019EvolvingMeta}
F.~D. Mesentier~Silva, R.~Canaan, S.~Lee, M.~C. Fontaine, J.~Togelius, and
  A.~K. Hoover, ``{Evolving the hearthstone meta},'' in {\em IEEE Conference on
  Computatonal Intelligence and Games, CIG}, vol.~2019-August, IEEE Computer
  Society, 8 2019.

\bibitem{Chen2018Q-DeckRec:Games}
Z.~Chen, C.~Amato, T.-H. Nguyen, S.~Cooper, Y.~Sun, and M.~S. El-Nasr,
  ``{Q-DeckRec: A Fast Deck Recommendation System for Collectible Card
  Games},'' {\em IEEE Conference on Computatonal Intelligence and Games, CIG},
  vol.~2018-August, 6 2018.

\bibitem{Bhatt2018ExploringSpace}
A.~Bhatt, S.~Lee, F.~De~Mesentier~Silva, C.~W. Watson, J.~Togelius, and A.~K.
  Hoover, ``{Exploring the hearthstone deck space},'' in {\em ACM International
  Conference Proceeding Series}, (New York, NY, USA), pp.~1--10, Association
  for Computing Machinery, 8 2018.

\bibitem{Mahlmann2012EvolvingDominion}
T.~Mahlmann, J.~Togelius, and G.~N. Yannakakis, ``{Evolving card sets towards
  balancing dominion},'' in {\em 2012 IEEE Congress on Evolutionary
  Computation}, pp.~1--8, IEEE, 6 2012.

\bibitem{Fontaine2019MappingBoundaries}
M.~C. Fontaine, S.~Lee, L.~B. Soros, F.~D.~M. Silva, J.~Togelius, and A.~K.
  Hoover, ``{Mapping Hearthstone Deck Spaces through MAP-Elites with Sliding
  Boundaries},'' in {\em Genetic and Evolutionary Computation Conference}, 4
  2019.

\bibitem{LucasHanke2017AGames}
L.~C. Lucas~Hanke, ``{A Recommender System for Hero Line-Ups in MOBA Games},''
  in {\em AAAI Conference on Artificial Intelligence and Interactive Digital
  Entertainment}, 2017.

\bibitem{Chen2018TheDrafting}
Z.~Chen, T.-H.~D. Nguyen, Y.~Xu, C.~Amato, S.~Cooper, Y.~Sun, and M.~S.
  El-Nasr, ``{The art of drafting},'' in {\em Proceedings of the 12th ACM
  Conference on Recommender Systems}, (New York, NY, USA), pp.~200--208,
  Association for Computing Machinery (ACM), 9 2018.

\bibitem{Summerville2016Draft-AnalysisLearning}
A.~Summerville, M.~Cook, and B.~Steenhuisen~Datdota, ``{Draft-Analysis of the
  Ancients: Predicting Draft Picks in DotA 2 Using Machine Learning},'' in {\em
  Twelfth Artificial Intelligence and Interactive Digital Entertainment
  Conference}, 9 2016.

\bibitem{Yu2019E-SportsNetwork}
C.~Yu, W.~n. Zhu, and Y.~m. Sun, ``{E-Sports Ban/Pick Prediction Based on
  Bi-LSTM Meta Learning Network},'' in {\em Lecture Notes in Computer Science
  (including subseries Lecture Notes in Artificial Intelligence and Lecture
  Notes in Bioinformatics)}, vol.~11632 LNCS, pp.~97--105, Springer Verlag, 7
  2019.

\bibitem{Matthews2012CompetingDomains}
T.~Matthews, S.~D. Ramchurn, and G.~Chalkiadakis, ``{Competing with Humans at
  Fantasy Football: Team Formation in Large Partially-Observable Domains},''
  {\em The Twenty-Sixth AAAI Conference on Artificial Intelligence}, 2012.

\bibitem{Datta2012CapacitatedNetworks}
S.~Datta, A.~Majumder, and K.~V. Naidu, ``{Capacitated team formation problem
  on social networks},'' in {\em Proceedings of the ACM SIGKDD International
  Conference on Knowledge Discovery and Data Mining}, vol.~12, (New York, New
  York, USA), pp.~1005--1013, ACM Press, 5 2012.

\bibitem{Gutierrez2016TheSociometry}
J.~H. Guti{\'{e}}rrez, C.~A. Astudillo, P.~Ballesteros-P{\'{e}}rez,
  D.~Mora-Meli{\`{a}}, and A.~Candia-V{\'{e}}jar, ``{The multiple team
  formation problem using sociometry},'' {\em Computers and Operations
  Research}, vol.~75, pp.~150--162, 11 2016.

\bibitem{Lappas2009FindingNetworks}
T.~Lappas, K.~Liu, and E.~Terzi, ``{Finding a team of experts in social
  networks},'' in {\em Proceedings of the ACM SIGKDD International Conference
  on Knowledge Discovery and Data Mining}, (New York, New York, USA),
  pp.~467--475, ACM Press, 2009.

\bibitem{Saxe2020BotDraftsim}
R.~Saxe, ``{Bot Drafting the Hard Way: A New Drafting Model for Archetypes and
  Staying Open - Draftsim},'' 2020.

\bibitem{CardsphereCardsphere:Simulator}
{Cardsphere}, ``{Cardsphere: MTG Draft Simulator}.''

\bibitem{MTGDraftKingMTGDraftKing:Gathering}
{MTGDraftKing}, ``{MTGDraftKing: draft and sealed simulator for Magic the
  Gathering}.''

\bibitem{WizardsoftheCoast2013Magic:Rules}
{Wizards of the Coast}, ``{Magic: The Gathering Comprehensive Rules},'' 2013.

\bibitem{Churchill2019Magic:Complete}
A.~Churchill, S.~Biderman, and A.~Herrick, ``{Magic: The Gathering is Turing
  Complete},'' {\em arXiv preprint arxiv:1904.09828}, 3 2019.

\bibitem{Chatterjee2016TheGathering}
K.~Chatterjee and R.~Ibsen-Jensen, ``{The Complexity of Deciding Legality of a
  Single Step of Magic: the Gathering},'' {\em 2016 European Conference on
  Artificial Intelligence (ECAI)}, 2016.

\bibitem{Rosewater2020ToGATHERING}
M.~Rosewater, ``{To the Death | MAGIC: THE GATHERING},'' 2020.

\end{thebibliography}
\end{document}